\documentclass{dsc}

\usepackage[latin1,utf8]{inputenc} 
\usepackage{amsmath}
\usepackage{amsfonts}
\usepackage{amssymb} 
\usepackage{lmodern}
\ifpdf \usepackage[pdftex]{graphicx} \pdfcompresslevel=9
\else \usepackage[dvips]{graphicx} \fi
\bibliography{reference}

\usepackage[hidelinks]{hyperref}
\usepackage{multirow}
\usepackage{xcolor}
\usepackage{adjustbox}
\usepackage{booktabs}
\usepackage{dblfloatfix}
\title[Frequency-splitting MCA]{Motion Cueing Algorithm for Effective Motion Perception: A frequency-splitting MPC Approach}

\author[Jain]
       {\textbf{Vishrut Jain$^{\text{1}}$,
       Andrea Lazcano$^\text{2}$,
       Riender Happee$^\text{1}$ and
       Barys Shyrokau$^\text{1}$}}
       
\begin{document}

\setcounter{page}{1}

\maketitle
\begin{affiliation}
{
(1) Delft University of Technology, Faculty of Mechanical, Maritime and Materials Engineering, Cognitive Robotics, 2628 CD, e-mail: \{V.J.Jain, R.Happee, B.Shyrokau\}@tudelft.nl\\
(2) Toyota Motor Europe, Zaventem, Belgium, e-mail: \ Andrea.Lazcano@toyota-europe.com
}
\end{affiliation}

\begin{abstract}
Model predictive control (MPC) is a promising technique for motion cueing in driving simulators, but its high computation time limits widespread real-time application. This paper proposes a hybrid algorithm that combines filter-based and MPC-based techniques to improve specific force tracking while reducing computation time. The proposed frequency-splitting algorithm divides the reference acceleration into low-frequency and high-frequency components. The high-frequency component is provided as a reference to the translational motion to avoid workspace limit violations, while the low-frequency component is for tilt coordination. 
The total acceleration serves as a reference for combined specific force with the highest priority to enable compensation of deviations from its reference values.
The algorithm uses constraints in the MPC formulation to account for workspace limits and workspace management is applied. These limit platform acceleration and velocity near workspace boundaries for better workspace utilization.
The investigated scenarios were a step signal, a multi-sine wave and a recorded real-drive slalom maneuver.
Based on the conducted simulations for 40 steps prediction horizon, the algorithm produces approximately 15\% smaller root means squared error (RMSE) for the step signal compared to the state-of-the-art. Around 16\% improvement is observed when the real-drive scenario is used as the simulation scenario, and for the multi-sine wave, an improvement of about 90\% is observed.
At higher prediction horizons the algorithm matches the performance of a state-of-the-art MPC-based motion cueing algorithm. Finally, for all prediction horizons, the frequency-splitting algorithm produced faster results.
The pre-generated references reduce the required prediction horizon and computational complexity while improving tracking performance. Hence, the proposed frequency-splitting algorithm outperforms state-of-the-art MPC-based algorithm and offers promise for real-time application in driving simulators.
\end{abstract}


\section*{Introduction}

Since the introduction of driving simulator technology, the primary goal has been to bridge the gap between the driving simulator experience and the real in-vehicle experience. Significant advancements have been achieved in this pursuit, although there is still room for improvement in accurately replicating the motion experience of being inside a vehicle.

To create a virtual environment with the aim of recreating the in-vehicle experience; visual, audio, vestibular and haptic cues are provided in the driving simulator \citep{shyrokau2018effect}.
However, the motion control of the simulator platform is challenging due to the limited workspace.

The motion cueing algorithm (MCA) implemented in the simulator aims to emulate the motion of a vehicle through various control strategies. The classical approach is to use tilt-coordination to recreate low-frequency accelerations, and high-frequency accelerations by linear accelerations of the platform \citep{Stratulat2011, Seehof2014}. 

A more enhanced approach is based on model predictive control \citep{Bruschetta2017, Khusro2020, Lamprecht2021}. However, this technique faces a drawback in terms of its high computation time, which poses a limitation for real-time applications.
To reduce computational costs, explicit MPC can be applied by pre-computing the solution and using it in the form of a look-up table. This method significantly reduces online computation time \citep{Fang2012}. A 2 DoF motion cueing is developed and extended by incorporating a vestibular model in \citep{Munir2017}. While this approach offers the advantage of reducing online computation time, it encounters challenges related to memory storage and limitations on utilizing large prediction horizons with fast sampling rates. The issue arises from the exponential growth in computation time for control region calculations as the problem's complexity and scope increase. As a result, there is a need for a more practical alternative to address these concerns. Therefore, a 4 DoF MCA is proposed using a combination of explicit (offline) and implicit (online) MPCs in \citep{chadha2023}. A 4 DoF explicit MPC is used to provide an initial educated guess to the implicit MPC, resulting in faster convergence.

This paper presents an approach to tackle the problem of computation time for an MPC-based motion cueing algorithm, at the same time improving the specific force tracking performance compared to the state-of-the-art MPC approaches.

The proposed algorithm merges the features of filter-based and MPC-based MCAs to improve the accuracy of tracking the reference specific force. Like filter-based algorithms, the reference signal is separated into low-frequency and high-frequency components, which are reproduced using tilt coordination and translational motion, respectively. To achieve optimal tracking of these references and the overall reference specific force while considering workspace limitations, an MPC approach is employed.

\begin{figure*}[h]
	\centering
	\includegraphics[width=0.8\linewidth]{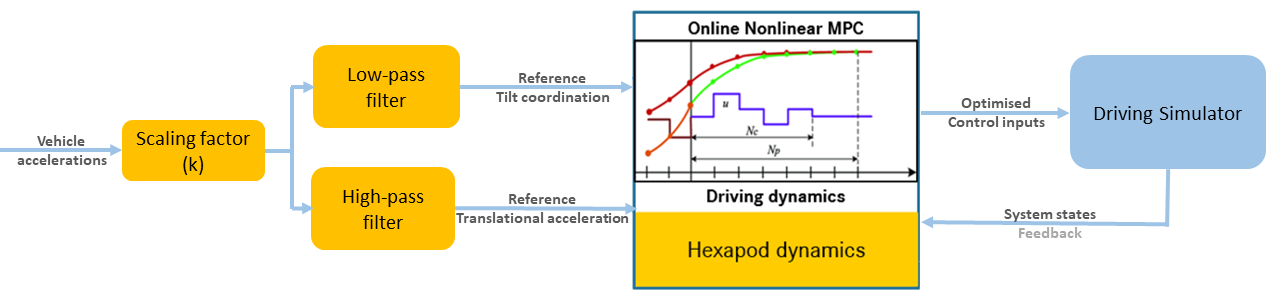}
	\caption{MPC Scheme for the frequency-splitting MCA}
	\label{fig:scheme}
\end{figure*}

\section*{Methodology}
The algorithm presented in this study utilizes a combination of filter-based and MPC-based MCA methods to improve the tracking performance of the reference specific force. The structure of the algorithm is shown in \autoref{fig:scheme}. The vehicle accelerations are scaled-down and passed through high-pass and low-pass filters. The high-frequency component obtained from the reference signal serves as a reference for the translational motion of the platform. This is because sustaining the low-frequency component could potentially result in violations of the workspace limits. The low-frequency component is used as a reference for tilt coordination. 
\subsection*{MCA formulation}
The cost function of an MPC minimises the squared error between the reference values and actual values of the outputs ($y_k$), inputs ($u_k$), and states ($x_k$) over a prediction horizon of $N$ future samples: 
\begin{eqnarray}\label{minimise}
& u_k = &\arg \min _{u_k}\frac{1}{N} \sum_{k=1}^N [ (x_k - \hat{x}_k)^T W_x (x_k - \hat{x}_k) \nonumber \\
&&+(y(x_k,u_k)-\hat{y}_k)^T W_y (y(x_k,u_k)-\hat{y}_k) \nonumber \\
&&+(u_k -\hat{u}_k)^T W_u (u_k -\hat{u}_k) ]
\end{eqnarray}

In MPC-based MCAs, it is common practice to penalise the difference between the simulator and vehicle motion outputs, i.e. specific force, as the output term.
The state term ($x_k -\hat{x}_k$) provides 'washout' to the platform by consistently trying to return it to its neutral position $\hat{x}_k$. The input term ($u_k -\hat{u}_k$) restricts high input values. In \autoref{minimise}, the tunable parameters are represented by weighting matrices $W_y,\ W_u$ and $W_x$.

In this work, the controlled degrees of freedom are longitudinal and lateral translation along with roll and pitch motion. Vertical and yaw motion are not part of the proposed MCA to reduce computational complexity. However, they can be added using a traditional washout filter. In this study, pitch and roll corresponding to vehicle motion are neglected due to small values. However, these can be directly added to the rotational inputs.

A triple integrator system is used to define the orientation and location of the platform. 

\begin{eqnarray}\label{triple_integrator}
    &&[\dot{s}_{hex},\ \dot{v}_{hex},\ \dot{a}_{hex},\ \dot{\theta}_{hex},\ \dot{\omega}_{hex},\ \dot{\alpha}_{hex}] \nonumber \\
     &&= [v_{hex},\ a_{hex},\ j_{hex},\ \omega_{hex},\ \alpha_{hex},\ j_{ang,hex}]
\end{eqnarray}
% \todo[inline]{change the representation, will save space}
where $s_{hex},\ v_{hex}$ and $a_{hex}$ represent the platform's translational position, velocity and acceleration respectively and $\theta_{hex},\ \omega_{hex}$ and $\alpha_{hex}$ represent the angular orientation, velocity and acceleration. $j_{hex}$ and $j_{ang,hex}$ represent the translational and angular jerk respectively which are the inputs to the system. 
The cost function penalizes jerks (inputs), limiting rapid changes in acceleration.\\
The specific force comprises of two components, arising through the translational motion and tilt-coordination. The translational component is the linear acceleration of the platform; the tilt component comes from the gravitational force subjected on the occupant of the simulator due to tilt (non-vertical orientation). The tilt component is described by:
\begin{equation}
    \dot{G} = -\omega_{hex} \times G
\end{equation}

The total specific force is defined as:
\begin{equation}
    f_{spec} = a_{hex}+G
\end{equation}

The output vector consists of the total specific force, tilt component of the specific force and translational acceleration of the platform ($y_{k}=[f_{spec}, G_{hex}, a_{hex}]$).

The pre-generated high-frequency and low-frequency components provide guidance for the tilt-coordination and translation motion of the simulator.
% This gives the algorithm combined benefits of classical washout and MPC-based algorithms. 

\subsection*{Scaling factor recommendation}
In this section, the scaling factor is designed, which reduces the reference platform motion relative to the vehicle motion.
In practice, a general trend is to use a scaling factor between 0.2 and 0.6 \citep{Lamprecht2021,bellem2017can}.
Subjective evaluations favoured such scaling in providing a more realistic motion perception.
The scaling factor also depends on the workspace limitations of the driving simulator being used.
In this work, it is assumed that the scenario is known prior to the simulation. The scaling factor is devised assuming complete information about the reference signal is available (considering the case of automated driving). 

Tilt-coordination is responsible for recreating sustained accelerations and a major portion of the specific force in the simulator. To ensure accurate recreation, the simulator's capability must meet or surpass the demand of the reference signal. Two recommended scaling factors were derived.

The first scaling factor is based on the maximum tilt angle. For this, we ensure that the reference specific force is always smaller than the platform's potential for generating specific force through tilt coordination.

\begin{eqnarray}
    &max|f_{spec,ref}|&\leq max|g sin (\theta_{tilt})|\\
    &k_{\theta} &= \frac{g sin(pi/6)}{max|f_{spec,ref}|}
\end{eqnarray}

'$\theta_{tilt}$' is limited to 30 deg corresponding to the minimum achievable '$cos(\theta_{tilt})$', the worst case scenario (wcs).

The second factor is based on the tilt-rate.

\begin{eqnarray}
    % & max |\dot{f}_{spec,ref}| &\leq |\dot{f}_{spec,tilt}|_{wcs}\nonumber \\
    % & max |\dot{f}_{spec,ref}| &\leq max |\frac{d g sin(\theta_{tilt})}{dt}| \nonumber \\
    & max |\dot{f}_{spec,ref}| &\leq  |\omega_{tilt} g cos(\theta_{tilt})|_{wcs} 
\end{eqnarray}

The '$\omega_{tilt}$' is restricted to a value of 3 deg/s to minimise the occurrence of false cues.

Hence the scaling factor can be given by:
\begin{equation}
    k_{\omega}\leq \frac{\pi}{60}\frac{g}{max \dot{f}_{spec,ref}} cos \frac{\pi}{6} 
\end{equation}

The smaller of these scaling factors can be chosen as the suggested scaling factor. However, larger scaling factors can also be utilized because, as explained below, the motion will be constrained within the available workspace through the combined implementation of workspace management techniques and the MPC algorithm.

\subsection*{Workspace management}
Due to the limited simulator workspace, proper utilisation is crucial, this subsection entails the steps taken to ensure better workspace utilisation.
\subsubsection*{Simulator capability limits}
Motion platform limits are added as constraints in the MPC formulation (\autoref{tab:limits}). 

\begin{table}[h]
	\caption{Applied motion limits due to rotation perception ($\omega_{perc}$) and simulator capability}
    \adjustbox{max width= \linewidth}{
    \begin{tabular}{l|lllll}
    \toprule
    Quantity & $\omega_{hex}$ & $\theta_{hex}$ & $v_{hex}$ & $a_{hex}$ & $s_{hex}$\\ \midrule 
    Limit    & $\pm 3 deg/s$     & $\pm 30 deg/s$    & $\pm 7.2 m/s$ & $\pm 9.81 m/s^2$ & $\pm 0.5 m$ \\
    % \bottomrule
    \end{tabular}}
	\label{tab:limits}
\end{table}

The tilt rate of the platform is restricted below the human perception threshold for angular velocities $\omega_{perc}$. Platform rotation below this threshold can be used to create the perception of translational acceleration. 
Additionally, braking constraints as proposed by \citet{Fang2012} are incorporated in the algorithm. As the workspace limits approach, the braking constraints limit the platform velocity and tilt rate, ensuring better workspace utilisation.
The relation for the constraint on displacement and angular orientation is:
\begin{eqnarray}
   s_{brk} = &s_{hex} +c_v v_{hex} T_{brk,s} + 0.5 c_{u} a_{hex} T^2_{brk,s}\\
    \theta_{brk} = &\theta_{hex} + c_{\omega} \omega_{hex} t_{brk,\theta} +0.5 c_{u} \alpha_{hex} T^2_{brk,\theta}
\end{eqnarray}

where $c_{v} = 1,\ c_{\omega} = 1, c_{u} = 0.45, T_{brk, \theta} = 0.5$, $T_{brk, s} = 2.5$.\\
The workspace management ensures that the limits of the platform are not reached, without any contribution to the cost function. 
As the platform moves in both longitudinal and lateral directions, the constraint is thus applied to the resultant displacement of the platform. The constraints on the displacement and tilt angle are:
\begin{eqnarray}
    -0.5 & \leq \sqrt{s_{brk,long}^2 +s_{brk,lat}^2} & \leq 0.5  \\
    -30 & \leq \theta_{brk} & \leq 30
\end{eqnarray}
\subsubsection*{Washout}
\begin{figure}[h]
    \centering
    \includegraphics[width = 0.45\textwidth]{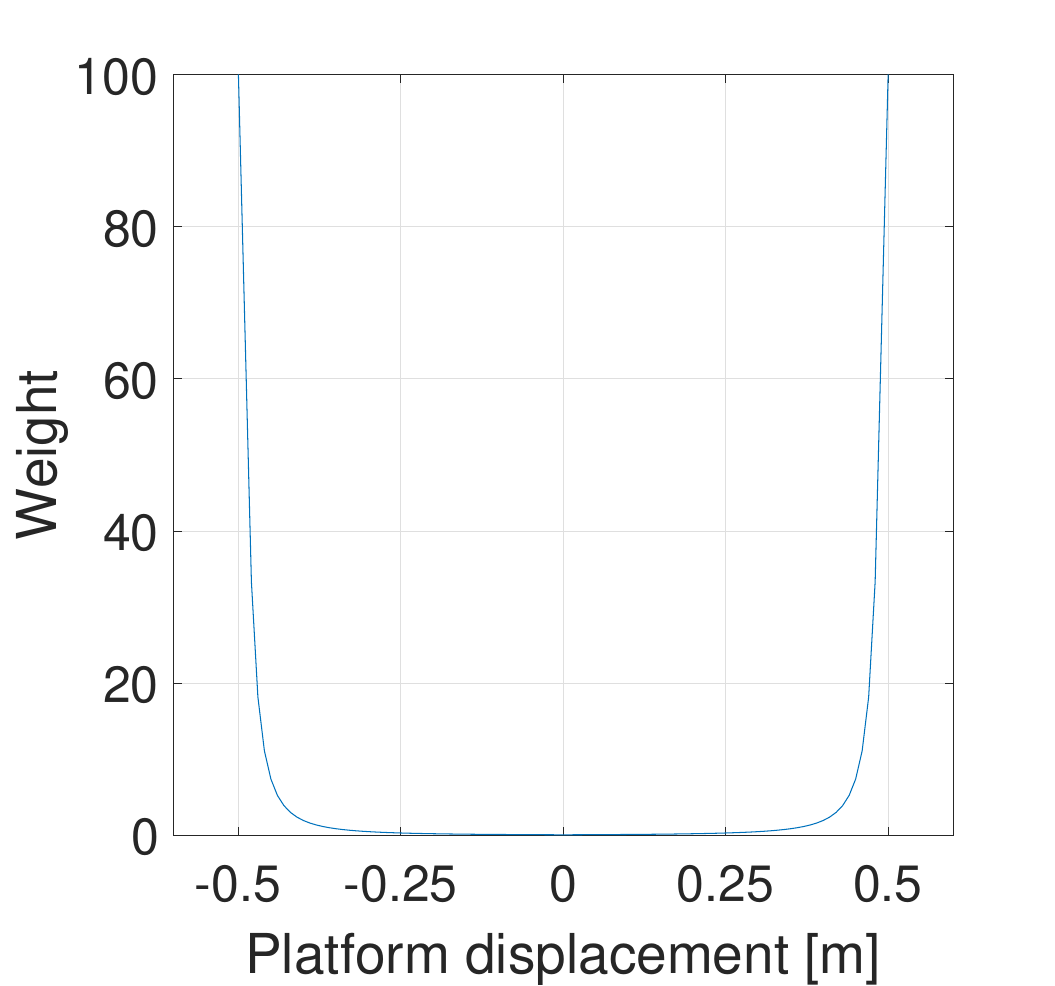}
    \caption{Adaptive position weight for the platform displacement (\autoref{adaptive_wt})}
    \label{fig:disp_wt}
\end{figure}

The simulator has the maximum potential of recreating any motion from its neutral position, away from the workspace limits. 
Thus, to ensure that the driving simulator performs at its maximum potential, the platform is ideally operating around its neutral position. 
This is performed by adding penalisation to the state term in \autoref{minimise} which brings the platform back to its neutral position ($\hat{x}_{k}$).\\
In this work, we use adaptive weights for the washout rather than constant weights.
The penalisation weight changes with the value of the state. This allows a single adaptive setting for all the scenarios, rather than tuning the washout weights for each scenario.\\
The relations for the adaptive weight are:
\begin{eqnarray} 
    &w_{s} &= \frac{k_1}{k_2*(|s_{hex}|-0.5)^2+\Delta} \label{adaptive_wt} \\ 
    &w_{\theta} &= \frac{k_3}{k_2*(|\theta_{hex}|-30*\pi/180)^2+\Delta}
\end{eqnarray}

where $k_{1}$, $k_{2}$ and $k_{3}$ are the parameters through which the shape of the weight function can be changed. $\Delta$ (here 0.01) is a small value added to the denominator to avoid singularity. The values selected for the simulations are $k_1 = 1$, $k_2 = 50$  and $k_3 = 0.1$. The variation of weight with the platform displacement can be seen in \autoref{fig:disp_wt}. The shape of this weight can be changed by tuning $k_1,k_2,k_3$ and $\Delta$. 

\subsection*{Simulation parameters}
The weights used in this study were determined through a manual tuning approach. 
They correspond to the best simulation outcomes for the real-drive scenario outlined in the subsection "Scenarios". 
The following weight settings are selected:
\begin{table}[h]
\centering
\caption{Selected values of the simulation parameters}
\adjustbox{max width= \linewidth}{
\begin{tabular}{c c}
\toprule
Parameter & Value \\
\midrule
Cut-off frequency & 0.5 Hz \\
weight on linear acceleration tracking $w_{a,hex}$ & 1 \\
weight on tilt-coordination tracking $w_{G}$ & 1 \\
weight on total specific force tracking $w_{f,spec}$ & 5 \\
weight on linear jerk $w_{j}$ & $10^{-2}$ \\
weight on angular jerk $w_{ang,j}$ & $10^{-3}$ \\
\bottomrule
\label{tab:constraints}
\end{tabular}}
\end{table}

Higher weight is provided to the specific force tracking, aiming to create realistic perceived motion. A lower weight is selected for linear acceleration and tilt-coordination whereby the MCA tries to follow these two references, but prioritises specific force flexibly using acceleration and tilt coordination. Small weights on jerk and angular jerk are provided to avoid oscillations in the specific force, due to rapidly changing acceleration. The weights for angular orientation ($w_{\theta}$) and displacement ($w_{s}$) are selected as described in the section "Washout". 

The same weight settings were maintained for both the benchmark and proposed algorithms, eliminating any potential biases. However, the proposed algorithm includes additional penalization components ($w_{G}$ and $w_{a,hex}$), addressing linear acceleration and tilt-coordination tracking separately.

\section*{Simulation environment}
The scenario and the setup for the simulations are described in this section. 
The optimisations were performed using the \verb|ACADO| \citep{houska2011acado} solver, to formulate the MPC. 
Maximum solver iterations are chosen to be 200, to ensure convergence and avoid sub-optimal solutions.
The optimisation has been performed on Intel(R) Xeon(R) W-2223 CPU @3.60GHz with 32GB RAM.

\subsection*{Benchmarking}
A state-of-the-art MPC-based MCA for specific force tracking is used as the benchmark for this study.
The algorithm still follows the structure defined in \autoref{minimise} with modification in the output term.
The output term becomes $y_{k,bench} = f_{spec}$ as specific force tracking is the sole objective of the algorithm.
The structure of the algorithm is adapted from \citep{van2020sensitivity}

\subsection*{Scenarios}
The cases considered for this paper are a step signal, a multi-sine wave and a recorded real-drive scenario.

\subsubsection*{Step signal}
The step signal gives a sudden increment in the reference specific force, to simulate an extreme dynamic maneuver. The step signal is defined by a rest period, then a consistent magnitude of $+0.8 m/s^2$ for 8 secs followed by a rest period again. This signal is provided as a reference in both longitudinal and lateral directions. This signal is also used to evaluate the tracking performance in \citep{Munir2017,chadha2023}.

\subsubsection*{Multisine}
This signal consists of 4 different sine waves, of frequency 0.1, 0.15, 0.2 and 0.5 Hz and amplitude 1, 0.8, 0.1 and 0.6 $m/s^2$ respectively. This is used to check the performance of the MCA for a continuously changing specific force. 
The different sine components ensure that different frequencies are covered in the same signal.

\subsubsection*{Real-drive scenario}
A slalom maneuver recorded during a real car driving experiment \citep{irmak2021objective} is also used for validation of the MCA.
The slalom with 4 $m/s^2$ at 0.2 Hz is one of the most aggressive maneuvers that a vehicle can be subjected to and was designed to elicit motion sickness in passive driving (i.e. as a user of an automated vehicle). The simulations focus on the initial 50 seconds of the maneuver.

\subsection*{Emulator simulation}
To evaluate the performance and computational costs, the software emulator has been used developed by the motion platform supplier E2M Technologies B.V. The multi-body modelling and the coordinate system are described in \citep{E2M_manu}.\\
This emulator represents the complex actual dynamics of the Delft Advanced Vehicle Simulator (DAVSi). The DAVSi is a 6 DoF driving simulator and using its emulator interface, tests can be performed without imparting any damage to the real system.\\
As the actual dynamics of the simulator differ from the predictions of the internal model in the MPC, the constraint on tilt-rate (\autoref{tab:constraints}) is violated in some instances. This creates instability in the algorithm. To tackle this instability the tilt rate is provided with a soft constraint, i.e. the violation of the constraint is allowed but penalised. \\
The constraint is reformulated as:
\begin{eqnarray}
    &-\omega_{th} &\leq \omega_{hex} + \delta \\
    & \omega_{hex} -\delta & \leq \omega_{th}\\
    & \delta & \geq 0
\end{eqnarray}
Where, $\omega_{th}$ is the threshold for tilt rate and the slack variable, $\delta$, is penalised in the cost function to reduce high violation of constraints. In this study, the penalisation for the delta, $w_{\delta}$ is chosen to be $10^5$.

\begin{figure*}[h]
    \centering
    \includegraphics[width = 0.7\textwidth]{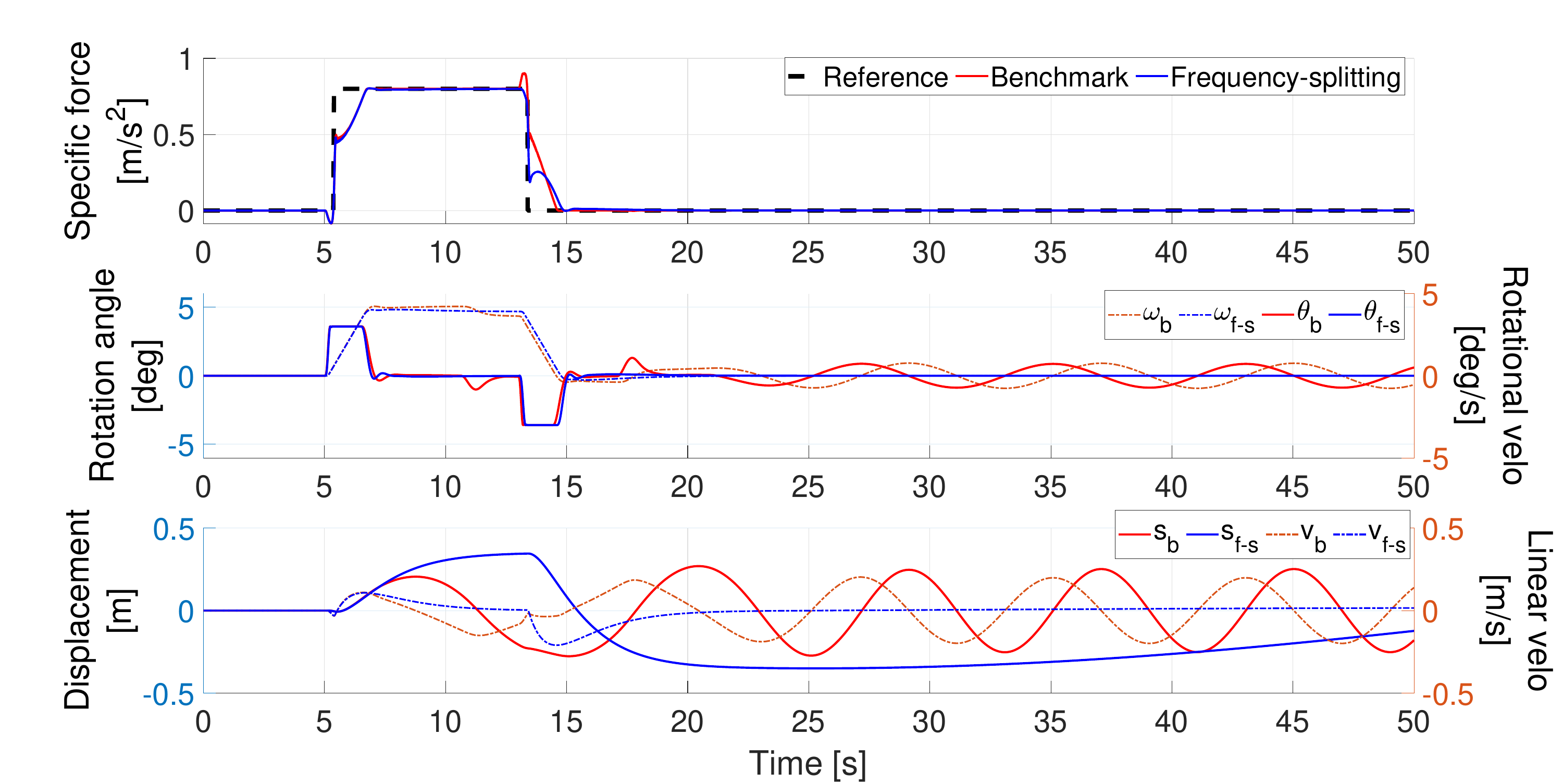}
    \caption{Specific force tracking for step signal: Longitudinal direction}
    \label{fig:step_long}
\end{figure*}

\section*{Results}

\subsection*{Prediction horizon analysis}
The simulations were performed at various prediction horizon lengths, with 0.01 s as the MPC time step, to evaluate the performance of the frequency-splitting algorithm compared to the benchmark algorithm. The obtained results are tabulated below. The signal used for this comparison was the real-drive scenario described in the above section. The scaling factor used for this analysis is 0.15 and the cut-off frequency for frequency-splitting is 0.5 Hz.

\begin{table*}[]
\caption{Performance of benchmark and frequency-splitting algorithms at different prediction horizons for the real driving scenario}
\label{table_ph}
\adjustbox{max width= \linewidth}{
\begin{tabular}{c|cccc|cccc}
\toprule
\multirow{2}{*}{\begin{tabular}[c]{@{}c@{}}Prediction horizon\\ (Steps)\end{tabular}} & \multicolumn{4}{c|}{Benchmark algorithm}              & \multicolumn{4}{c}{Frequency-splitting algorithm} \\    \cmidrule{2-9}
  & RMSE long & RMSE lat & RMSE total & Computation time & RMSE long & RMSE lat & RMSE total & Computation time \\
\midrule
25                                                                                    & 0.0295    & 0.1037   & 0.1078     & \textcolor{green}{18.2 s}           & 0.0186    & 0.0599   & 0.0627     & \textcolor{green}{18.5 s}           \\
30                                                                                    & 0.0096    & 0.0619   & 0.0626     & \textcolor{red}{30.7 s}           & 0.0179    & 0.0575   & 0.0602     & \textcolor{green}{28.3 s}           \\
35                                                                                    & 0.0075    & 0.0526   & 0.0531     & \textcolor{red}{48.6 s}           & 0.0174    & 0.0432   & 0.0466     & \textcolor{red}{39.2 s}           \\
40                                                                                    & 0.0075    & 0.0540   & 0.0545     & \textcolor{red}{67.0 s}           & 0.0170    & 0.0426   & 0.0459     & \textcolor{red}{48.0 s}           \\
50                                                                                    & 0.0067    & 0.0465   & 0.0470     & \textcolor{red}{128.6 s}          & 0.0152    & 0.0404   & 0.0432     & \textcolor{red}{68.0 s}           \\
100                                                                                   & 0.0064    & 0.0298   & 0.0305     & \textcolor{red}{1495.8 s}         & 0.0073    & 0.0263   & 0.0273     & \textcolor{red}{1106.7 s}         \\
150                                                                                   & 0.0064    & 0.0147   & 0.0160     & \textcolor{red}{4690.0 s}         & 0.0062    & 0.0168   & 0.0179     & \textcolor{red}{2885.9 s}        
\end{tabular}}
\end{table*}

From \autoref{table_ph}, it can be observed that up to a prediction horizon of 100 steps, the proposed algorithm outperforms the benchmark algorithm with a reduced total RMSE. In addition, the algorithm renders a faster convergence to the solution, with medium to long prediction horizons. The numbers highlighted in green correspond to real-time feasible and the red ones correspond to real-time infeasible solutions with the current implementation. \\ Hence, considering both accuracy and speed, at low prediction horizons, the proposed algorithm is clearly a better choice for motion cueing. However, at higher prediction horizons the benchmark algorithm outperforms the proposed algorithm. \\
It should be noted that even at higher prediction horizons, the frequency-splitting algorithm produces comparable results to the benchmark, providing a faster convergence to the solution. \\
In the remainder of this paper, a prediction horizon of 40 steps ($0.4s$) was chosen for both algorithms.

\subsection*{Specific force tracking performance}

\subsubsection*{Step signal}
In this simulation, the step signal is provided as the reference in both lateral and longitudinal directions. 
As the algorithm exhibits an identical response in both longitudinal and lateral directions, only the results pertaining to the longitudinal direction are presented in \autoref{fig:step_long}.
In the figures, the subscript $b$ in $\omega_{b},\theta_{b},s_{b}$ and $v_{b}$ represents simulations using the benchmark algorithm. Similarly, the subscript $f-s$ represents simulations using the proposed algorithm.\\
It can be seen as the platform is restricted in terms of tilt rate, at 5 s '$\omega$' reaches its threshold.
At this point, however, '$\theta$' is below its maximum limit.
Thus, the algorithms can not recreate the specific force at 5 s. However, it slowly converges to the reference, keeping a constant maximum allowable tilt rate. \\
The used scaling factor (k=1) is much larger than the designed scaling factor (k=0.11) in this case. 
Both algorithms show a similar response for the step input. 
The obtained RMSE for the specific force tracking in the case of the benchmark algorithm is 0.0657 $m/s^2$, and that for the frequency-splitting algorithm is 0.0557 $m/s^2$. This indicates a $15\%$ reduction in the specific force error.
In \autoref{fig:step_long}, the benchmark algorithm exhibits oscillations in rotation and displacement during the rest period after the step signal. 
The specific force is kept at 0 by the opposing magnitudes of the tilt component and translational component. 
The oscillations with the benchmark are reduced when increasing the horizon.
Using a horizon of 150 steps eliminates the oscillations, but this comes at a cost of approximately 70x increase in the computation time.
One approach to reducing oscillations is to increase the penalty on the control inputs. However, this comes at the expense of diminished tracking performance.

When the scaling factor is reduced to 0.5 and 0.11 the frequency-splitting algorithm shows an improvement of 35\% and 4\% respectively in RMSE when compared to the benchmark. The benchmark algorithm shows oscillations also with these reduced scaling factors.

\subsubsection*{Multi-sine}
\begin{figure}[h]
    \centering
    \includegraphics[width = 0.5\textwidth]{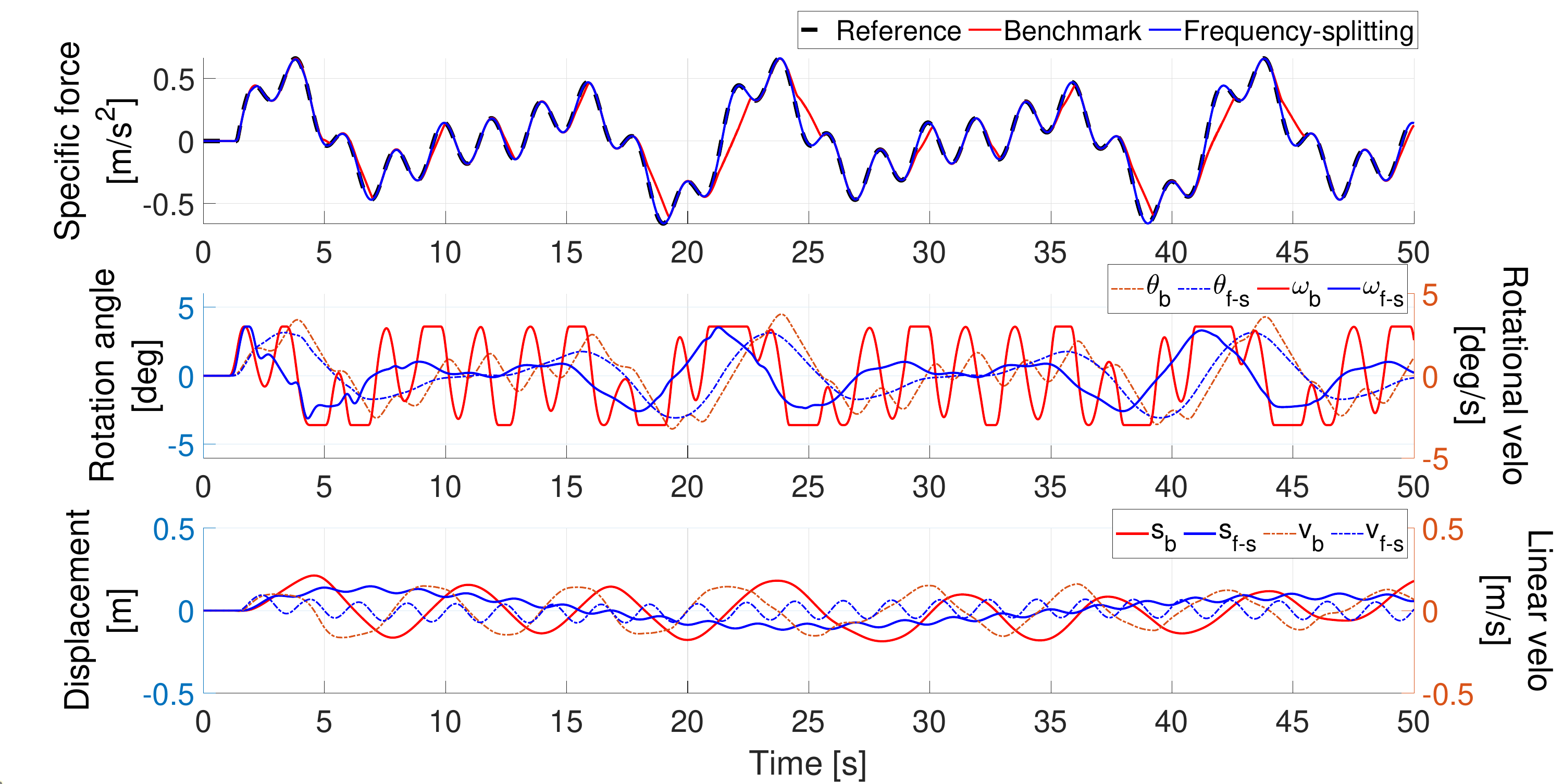}
    \caption{Simulation of the multi-sine wave: Longitudinal direction}
    \label{fig:multi_long}
\end{figure}

Using the motion scaling factor design, the obtained scaling factor is 0.295. 
For the simulation, we use a scaling factor of 0.3.
Due to identical responses in the longitudinal and lateral direction, only the results corresponding to the longitudinal direction are presented in \autoref{fig:multi_long}. 
The figure presents the comparison of specific force tracking between the benchmark and frequency-splitting algorithms. 
It can be seen that using the frequency-splitting algorithm, almost perfect tracking is obtained, whereas, the benchmark algorithm loses tracking at several places. 
It can also be observed that the benchmark algorithm reaches the perception threshold for the angular velocity at various regions prematurely and loses tracking.
However, in the case of the frequency-splitting algorithm, the references for tilt-coordination and linear acceleration ensure better workspace utilisation, resulting in better tracking performance.
The RMSE for the benchmark algorithm was found to be 0.0882 $m/s^2$, while for the frequency-splitting algorithm, it was significantly lower at 0.0085 $m/s^2$. This entails a reduction in the error by approximately $90\%$.

\subsubsection*{Real-drive scenario}

\begin{figure}[h]
    \centering
    \includegraphics[width = 0.5\textwidth]{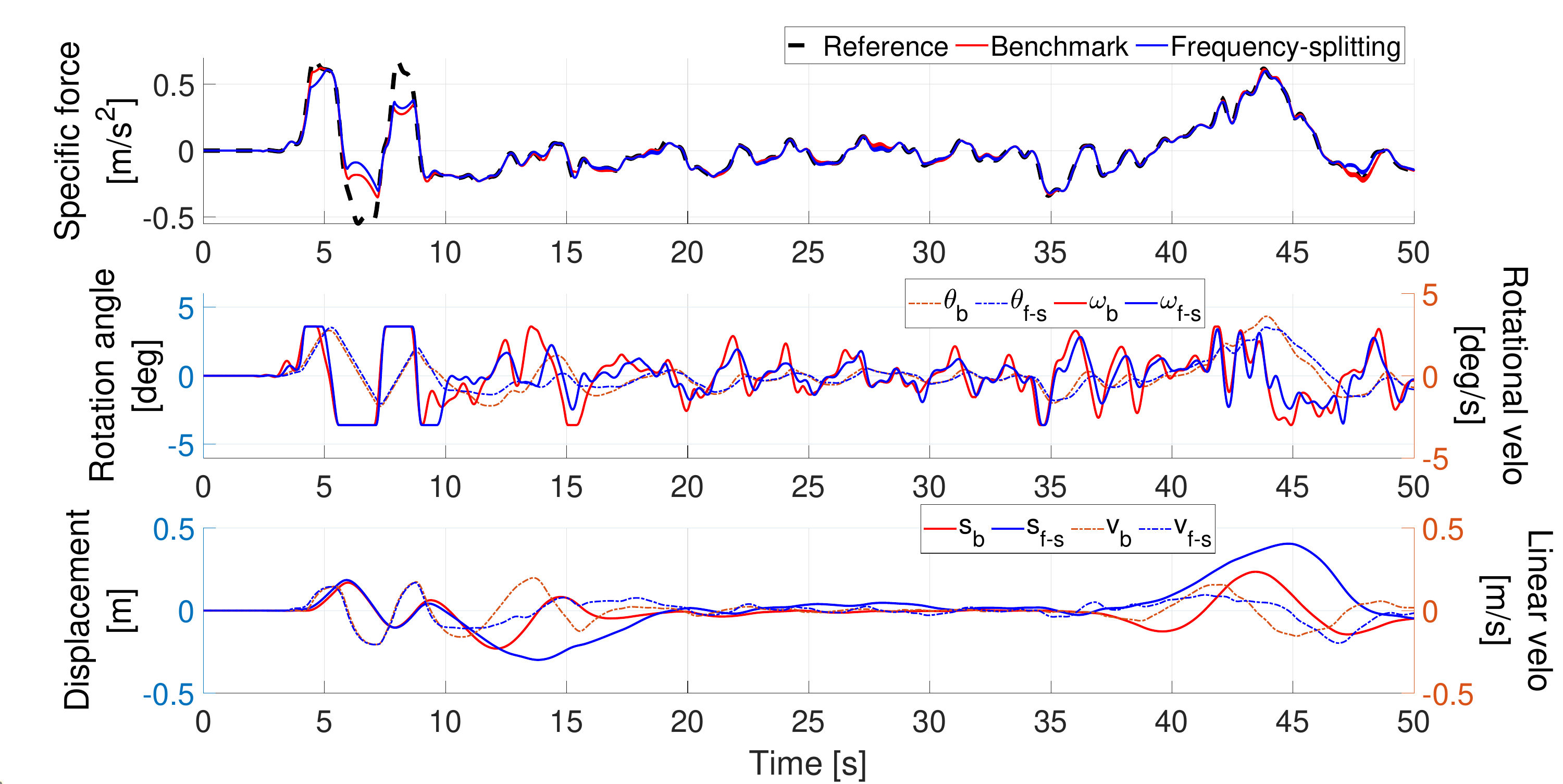}
    \caption{Simulation of the real-drive scenario: Longitudinal direction}
    \label{fig:slalom_long}
\end{figure}

\begin{figure}[h]
    \centering
    \includegraphics[width = 0.5\textwidth]{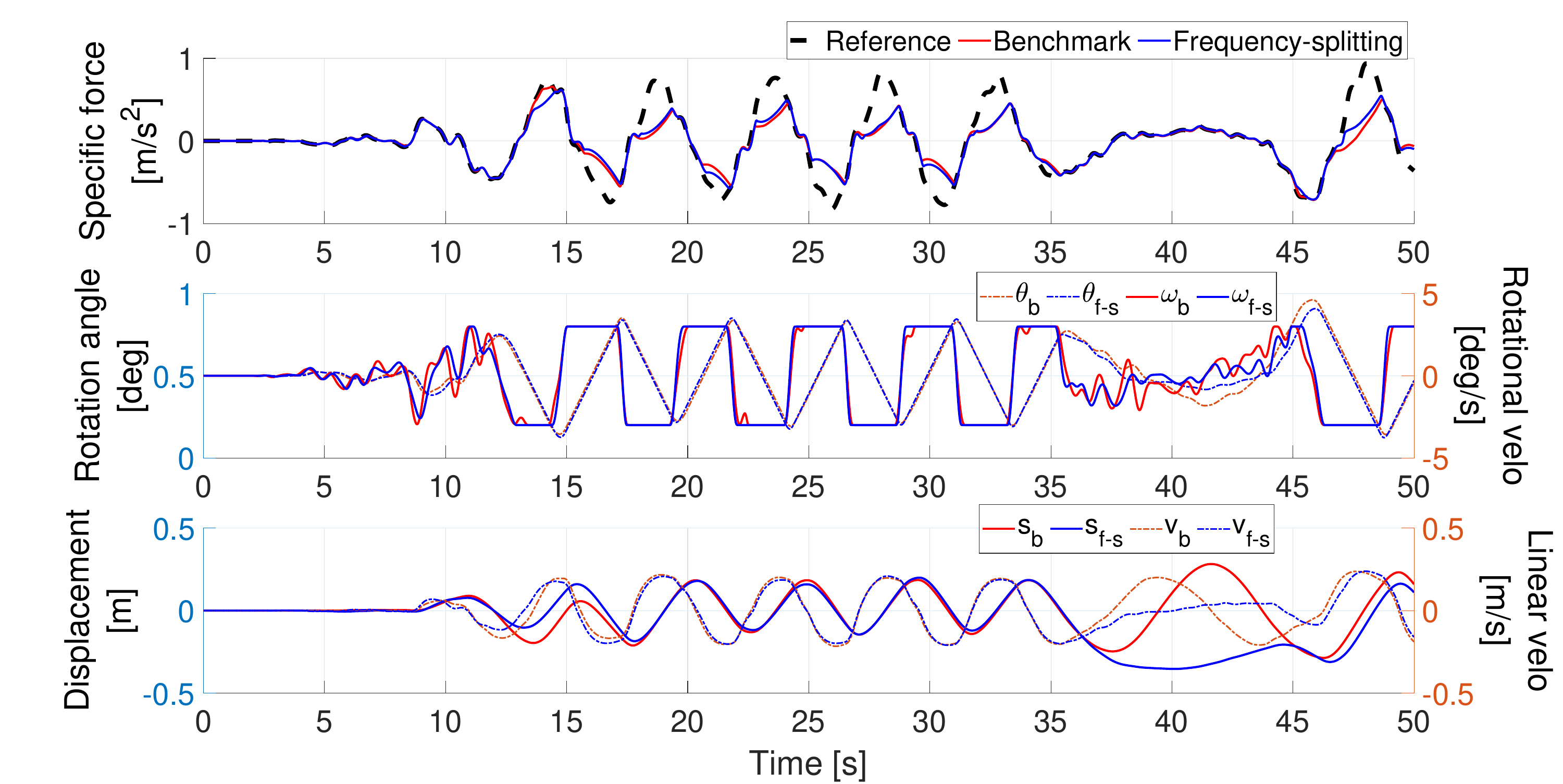}
    \caption{Simulation of the real-drive scenario: Lateral direction}
    \label{fig:slalom_lat}
\end{figure}

The simulation corresponding to the real-drive scenario is presented in \autoref{fig:slalom_long} for longitudinal and \autoref{fig:slalom_lat} for lateral motion.

Both algorithms show a similar response for the simulations and fail to follow the reference acceleration. In this case, the used scaling factor (0.2) is larger than the designed scaling factor (0.13), thus the driving simulator does not have the capability to recreate the reference accelerations. However, the frequency-splitting algorithm still produces slightly better tracking. The similarity in the performance is attributed to the heaviest penalisation provided to the specific force tracking.
The RMSE for the benchmark algorithm was found to be 0.2165 $m/s^2$, while for the frequency-splitting algorithm, it was 0.2031 $m/s^2$, indicating a reduction in the tracking error by approximately $7\%$.

\subsection*{Emulator simulations}
\begin{figure}[h]
    \centering
    \includegraphics[width = 0.5\textwidth]{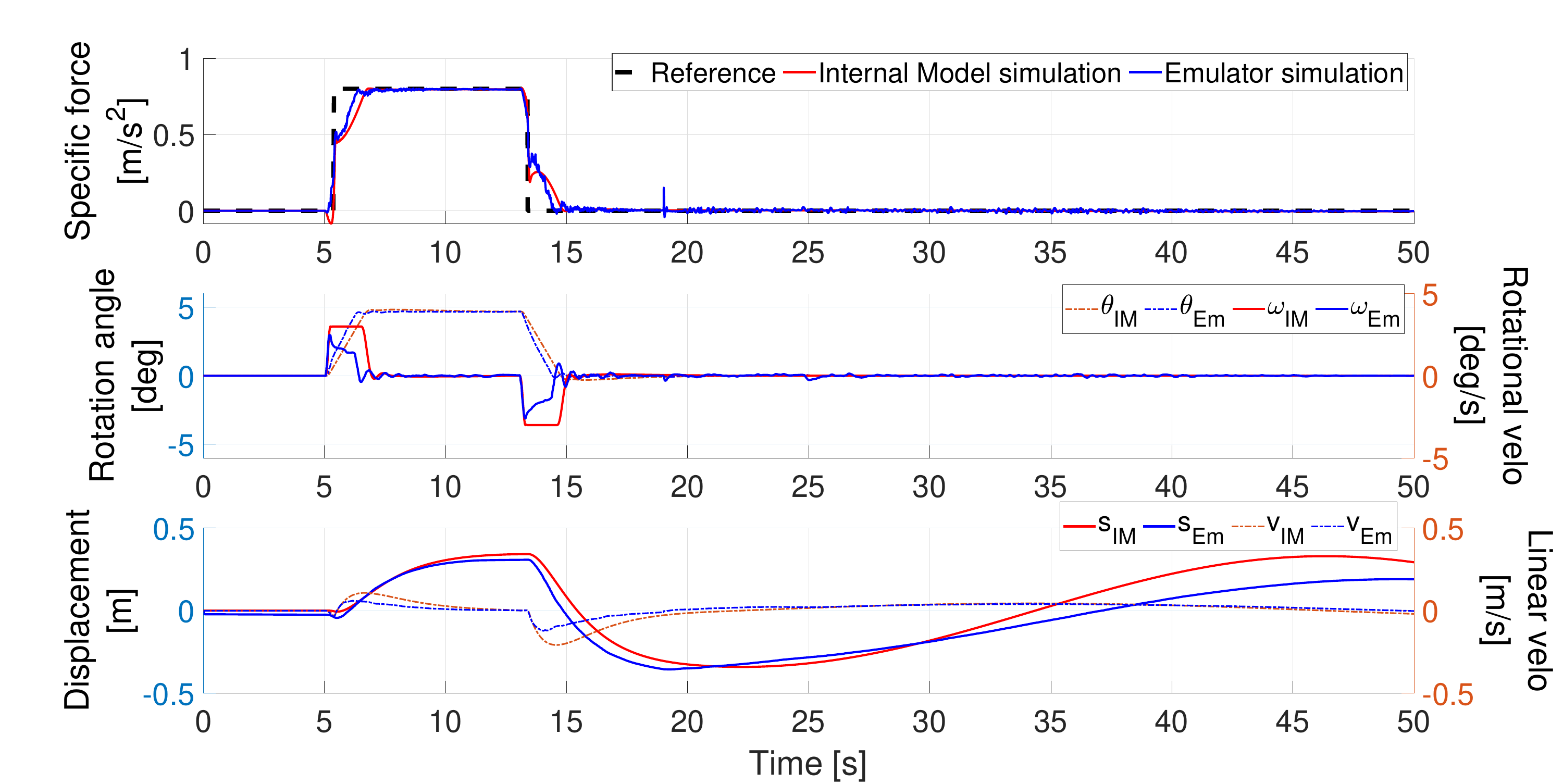}
    \caption{Simulation of the step signal via Emulator: Longitudinal direction}
    \label{fig:emul_step}
\end{figure}

\begin{figure}[h]
    \centering
    \includegraphics[width = 0.5\textwidth]{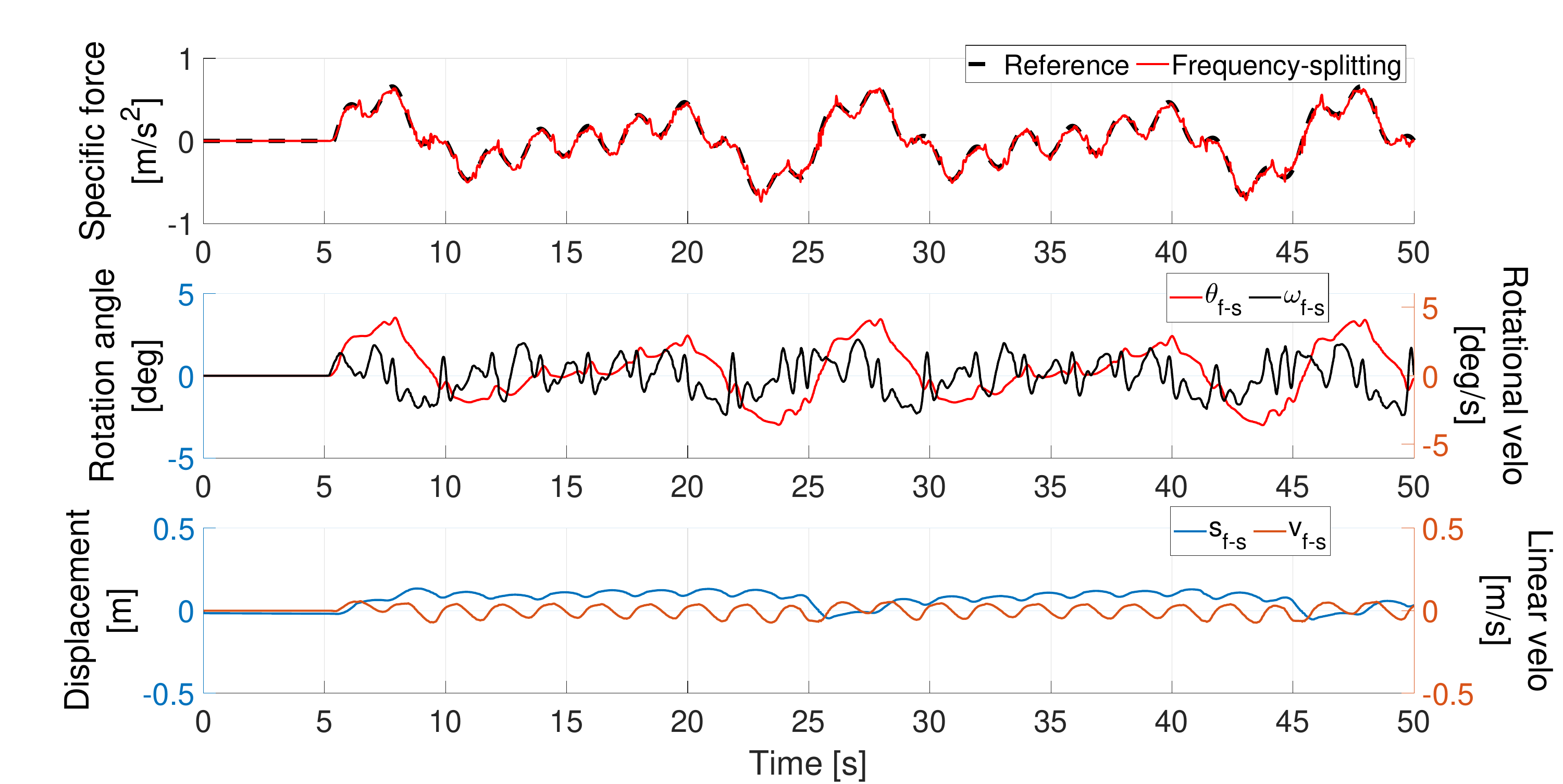}
    \caption{Simulation of the multi-sine signal via Emulator: Longitudinal direction}
    \label{fig:emul_multisine}
\end{figure}

\begin{figure}[h]
    \centering
    \includegraphics[width = 0.5\textwidth]{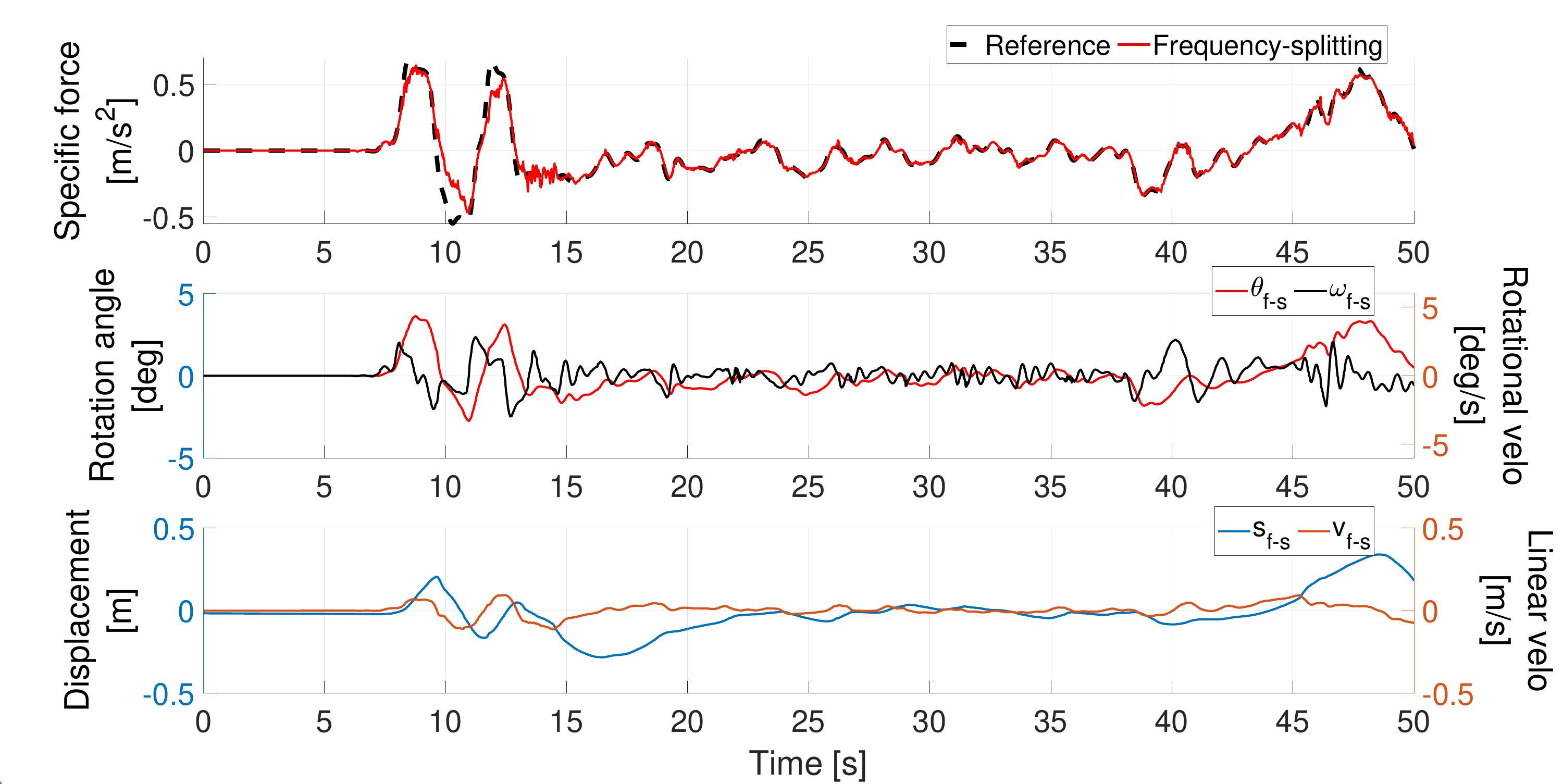}
    \caption{Simulation of real-drive scenario via Emulator: Longitudinal direction}
    \label{fig:emul_slalom_long}
\end{figure}

\begin{figure}[h]
    \centering
    \includegraphics[width = 0.5\textwidth]{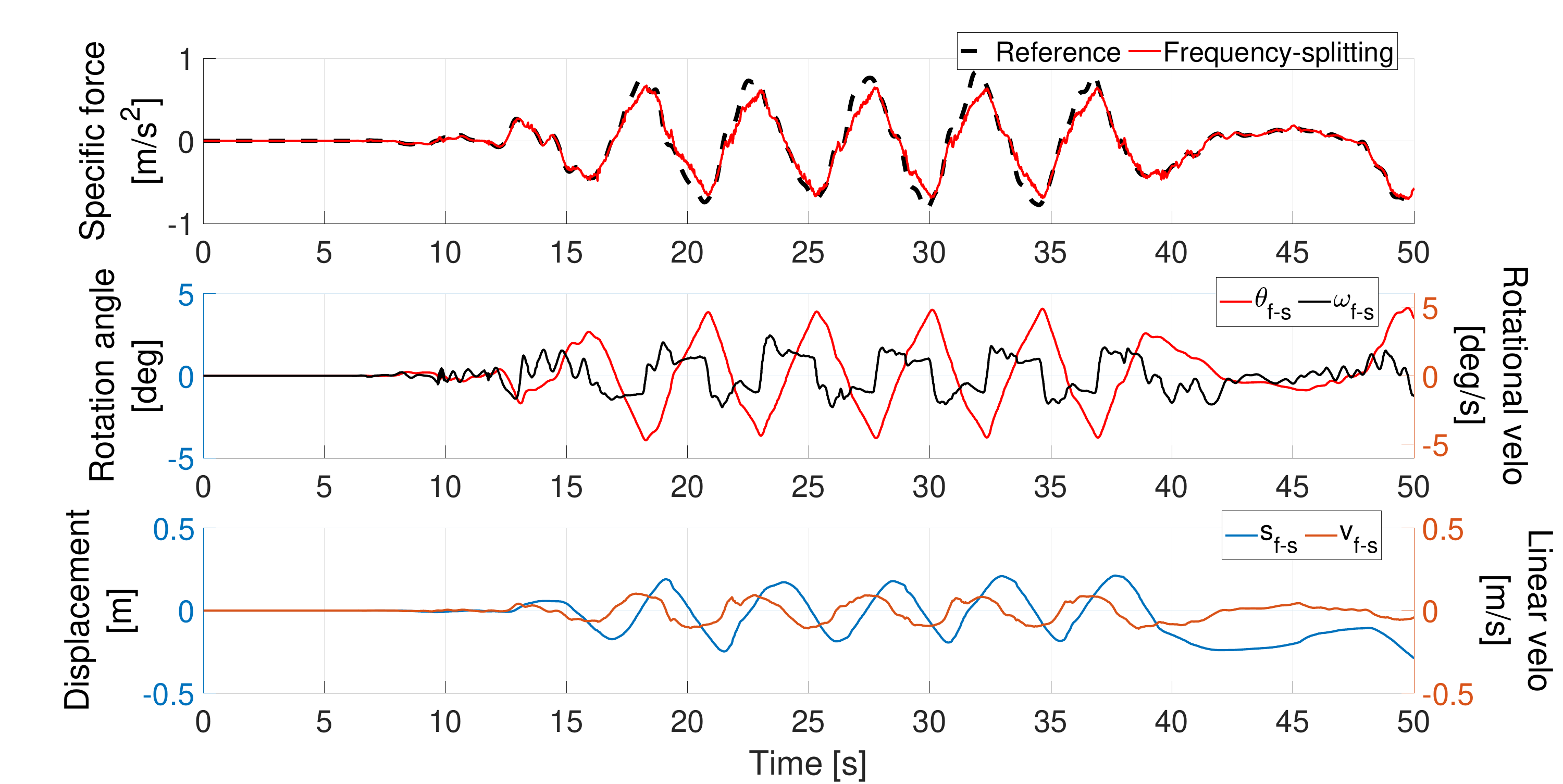}
    \caption{Simulation of the real-drive scenario via Emulator: Lateral direction}
    \label{fig:emul_slalom_lat}
\end{figure}

The algorithm is also simulated using the emulator for the driving simulator, to check the working of the algorithm in an actual driving simulator setting, taking into account the complex simulator dynamics. Due to the identical response for lateral and longitudinal directions in the case of step signal and multi-sine wave, only longitudinal responses are presented.
A simulation of the step signal using the frequency-splitting algorithm is presented in \autoref{fig:emul_step}. 
The figure showcases the results obtained using two different plant models: the blue plot depicts the results obtained using the emulator as a plant model, while the red plot shows the results obtained using the triple integrator system (defined in \autoref{triple_integrator}). The subscript used to represent the emulator as a plant is $Em$ and that for the internal model or triple integrator system is $IM$.
The figure depicts that the algorithm's behaviour using the emulator as the plant deviates marginally from its behaviour when the triple integrator system is used. Some oscillations are observed in the specific force around the reference. This is an expected behaviour as the internal model does not precisely match the simulator dynamics.
% % 
The emulator simulation for the multi-sine wave is presented in \autoref{fig:emul_multisine}. 
It can be seen that just like in \autoref{fig:multi_long} the frequency-splitting algorithm is able to provide satisfactory tracking of the reference multi-sine signal. Similar to the case of step signal emulator simulation, the specific force displayed oscillations around the reference. However, the magnitude of these oscillations is small (order of $10^{-3}\ m/s^2$) which is an acceptable deviation from the reference and should create a similar motion perception in the occupant of the driving simulator.
% %
The emulator simulation for the real-drive scenario for longitudinal direction is presented in \autoref{fig:emul_slalom_long} and that for the lateral direction is presented in \autoref{fig:emul_slalom_lat}. It can be observed that the simulations provide satisfactory tracking of the reference signal in both longitudinal and lateral directions.


Hence, the algorithm provides a satisfactory tracking of the reference signal in the emulator in the three scenarios studied in this paper. This indicates the algorithm is implementable in the driving simulator.
% \clearpage
\section*{Discussion}
The proposed algorithm provides pre-generated references for the platform's movements, utilizing the high-frequency and low-frequency components. This enables the algorithm to converge to a solution more quickly and effectively. In contrast, the state-of-the-art MPC algorithm provides limited guidance for the platform motion. This leads to additional movements as multiple configurations of tilt-coordination and linear acceleration can produce the same specific force. As a result,  the proposed algorithm demonstrates improved utilization of the workspace compared to the benchmark algorithm, particularly with shorter prediction horizons.

MPC with a longer prediction horizon can provide better performance within the simulator's workspace limits. However, computational cost increases and hinders real-time capability. Therefore, MPC with a shorter prediction horizon is used, and the platform could still reach its workspace limits. 

To tackle this issue, washout is introduced with a non-linear cost on platform position complemented with braking constraints. In the case of washout without braking constraints, the platform generates a sudden change in acceleration near the workspace limits resulting in false cues. Combined with braking constraints, the acceleration and velocity of the platform are restricted, and the workspace limits are never reached. As the result, there are no such sudden changes reducing the likelihood of false cues.

At high prediction horizons, the benchmark algorithm is observed to outperform the frequency-splitting algorithm marginally. Possibly due to the fact that at higher prediction horizons the MPC can produce near-optimal results. In the case of the frequency-splitting algorithm, there are multiple references. Frequency filters define the references for tilt-coordination and linear acceleration, which guide the algorithm for short prediction horizons but a suboptimal solution may be obtained.

According to \autoref{table_ph}, the current CPU enables the algorithm to achieve real-time capability for a prediction horizon of 30 steps (0.3 s). Moreover, even at higher prediction horizons, the proposed algorithm generates solutions faster than the benchmark. A preliminary implementation of the algorithm was conducted using dSPACE rapid prototyping hardware. This implementation involved a sampling time of 0.02 seconds and a prediction horizon of 0.8 seconds (40 steps). The results were obtained in hard real-time and demonstrated performance comparable to the findings presented in the paper.

The performance of the algorithm also depends on the selection of the frequency-splitting cut-off frequency. 
Such a cut-off frequency is also applied in classical washout algorithms. We selected 0.5 Hz which is of similar order as 0.4 Hz, 0.6 Hz \citep{asadi2015incorporating} and 1 Hz \citep{tajima2006driving}. We also explored values from 0.3-0.6 Hz resulting in similar results for the specific force. This however as expected changes the contributions coming from the linear movements and the tilt-coordination. For future study, the selection of cut-off frequency could be based on a power spectral analysis of the reference signals, e.g. selecting the cut-off frequency to achieve equal power distribution for linear motion and tilt-coordination. 

To evaluate the real-time feasibility and applicability of the proposed MCA, simulations were conducted with the emulator of the motion platform. The emulator motion corresponds to the commanded behaviour and only marginally deviated due to more complex modelled dynamics.

\section*{Conclusion}
In this study, a frequency-splitting MCA is proposed combining elements of a filter-based and MPC-based MCA to achieve better specific force tracking performance with faster convergence.

The proposed algorithm demonstrates a superior specific force tracking performance compared to the state-of-the-art MPC-based MCA for real-time capable prediction horizons. 
The frequency-splitting algorithm produces at least 15\% smaller error for the simulated scenarios when compared to the state-of-the-art. The results also show a significant 40\% enhancement in convergence speed when utilizing the proposed frequency-splitting algorithm.
The improved performance is attributed to the use of tilt-coordination and linear acceleration references in the frequency-splitting algorithm, which aid in better workspace utilization and faster convergence towards the desired specific force.

\section*{Limitations and future work}

The proposed frequency-splitting algorithm offers a promising solution for real-time specific force tracking but has a few limitations.
The algorithm requires prior knowledge of the entire drive. The scaling factor can be designed if the maneuver's aggressiveness is known since it depends on the maximum rate change of the specific force. Thus, a sudden change in the specific force can result in a very low scaling factor. To address this issue, we suggest limiting the scaling factor to a minimum value.

In this study, the cut-off frequency was manually tuned to determine the algorithm's performance. However, an adaptive approach or using power spectral analysis to determine the cut-off frequency can result in a better performance.

Although the algorithm demonstrated real-time capability through simulations and analysis, no hard real-time implementation in a driving simulator was conducted. Future work focuses on human-in-the-loop experiments to evaluate the algorithm's effectiveness.

\section*{Acknowledgement}
We thank Dr. Irmak for sharing the experimental data. This work was funded by Toyota Motor Europe.

\printbibliography

\end{document}